\documentclass[letterpaper, 10 pt, conference]{ieeeconf}  
\usepackage{times}
\usepackage[scientific-notation=true]{siunitx}

\IEEEoverridecommandlockouts                              

\usepackage{bm}
\newcommand{\citet}[1]{\cite{#1}}
\newcommand{\citep}[1]{\cite{#1}}

\usepackage{multicol}
\usepackage{balance}
\usepackage[bookmarks=true]{hyperref}

\usepackage[utf8]{inputenc} 
\usepackage[T1]{fontenc}    
\usepackage{hyperref}       
\usepackage{url}            
\usepackage{booktabs}       
\usepackage{amsfonts}       
\usepackage{nicefrac}       
\usepackage{microtype}      
\usepackage{lipsum}

\usepackage{amssymb}
\usepackage{caption}
\usepackage{subcaption}

\usepackage{graphicx}
\usepackage{xcolor}
\usepackage{siunitx}
\usepackage{amsmath}
\newcommand{\sect}[1]{Sec.~\ref{#1}}
\newcommand{\fig}[1]{Fig.~\ref{#1}}
\newcommand{\eq}[1]{Eq.~\eqref{#1}}

\newcommand{\tab}[1]{Tab.~\ref{#1}}

\usepackage{float}

\usepackage{siunitx}

\newcommand{\code}[0]{\mbox{\small{\url{https://github.com/PisterLab/BotNet}}}}
\newcommand{\simname}[0]{BotNet}

\begin{document}

\title{\LARGE \bf \simname: A Simulator for Studying the Effects of Accurate Communication Models on Multi-agent and Swarm Control}

\author{Mark Selden\authorrefmark{1}, Jason Zhou\authorrefmark{1}, Felipe Campos\authorrefmark{1}, Nathan Lambert\authorrefmark{1},  Daniel Drew\authorrefmark{3}, Kristofer S. J. Pister\authorrefmark{1}
\thanks{Corresponding author: Nathan Lambert, \tt \href{mailto:nol@berkeley.edu}{nol@berkeley.edu}.}
\thanks{\authorrefmark{1}Department of Electrical Engineering and Computer Sciences, University of California, Berkeley, CA, USA.}%
\thanks{\authorrefmark{3}Department of Electrical and Computer Engineering, University of Utah, UT, USA.
}%
}


\maketitle

\begin{abstract}
Decentralized control in multi-robot systems is dependent on accurate and reliable communication between agents.
Important communication factors, such as latency and packet delivery ratio, are strong functions of the number of agents in the network.
Findings from studies of mobile and high node-count radio-frequency (RF) mesh networks have only been transferred to the domain of multi-robot systems to a limited extent, and typical multi-agent robotic simulators often depend on simple propagation models that do not reflect the behavior of realistic RF networks. 
In this paper, we present a new open source swarm robotics simulator, \simname,  with an embedded standards-compliant time-synchronized channel hopping (6TiSCH) RF mesh network simulator.
Using this simulator we show how more accurate communications models can limit even simple multi-robot control tasks such as flocking and formation control, with agent counts ranging from 10 up to 2500 agents.
The experimental results are used to motivate changes to the inter-robot communication propagation models and other networking components currently used in practice in order to bridge the sim-to-real gap. 
\end{abstract}

\section{Introduction}
\label{sec:intro}
Multi-agent systems are becoming more prevalent and interconnected with the emergence of low-cost radios and capable robots~\cite{brambilla2013swarm,floreano2015science}. 
As agent count increases, 
the classification of the research domain transitions from that of multi-robot systems, up to about 10 agents, to one of swarm robotics, with 10s, 100s, or even 1000s of agents~\cite{csahin2004swarm}.
Historically, the most successful method for control with lower agent counts has been centralized planning and optimization, with actions sent to all agents simultaneously~\cite{egerstedt2001formation,oh2015survey}. 
Imperfect sharing of information between agents in widely distributed systems and the significant computation requirements for planning with high agent counts are two major barriers to centralized control approaches at the swarm level.
Decentralized control, where decisions are made locally per-agent by communicating with one's neighbors~\cite{siljak2011decentralized}, is a common solution because it is more readily scalable to higher agent counts.
The success of decentralized control, however, relies on the communication between an agent and all of its neighbors, the success of which in turn depends heavily on the number and density of agents, data transfer rate, and communication network structure -- elements that are not typically studied at the scale of swarms nor with the dynamical complexity of mobile robots.
In this work, we introduce a tool, \simname, for studying the effects of local radio frequency (RF) communication on multi-agent control, and use it to demonstrate how the challenges of realistic communication can manifest during example tasks.

RF communication is a popular solution within the realm of decentralized control due to its relatively high information transmission rate, few restrictions on environment, existing commercially-available communication hardware, and more~\cite{yick2008wireless}.
RF networks have been studied extensively, especially in the context of stationary wireless sensor networks (WSNs)~\cite{akyildiz2010wireless}, but directly translating findings for use in mobile systems has proven difficult. 
Communication has proven to be a limiting factor for deployment of novel multi-agent controllers, as the dynamics of mesh networks can be difficult to predict and change rapidly~\cite{murphy_disaster_2016,shakeri_design_2019, drew2021multi}. 
Challenges to network reliability can come from environmental disturbances, such as fixed or mobile obstacles, task related challenges, such as large coverage areas, and from network-specific limitations, such as transmission collisions. 
For example, in the recent DARPA Subterranean Challenge~\cite{rouvcek2019darpa}, RF communications were extremely unreliable in underground environments. 
In disaster relief scenarios, the ad hoc wireless networks established by first responders can quickly become saturated and lead to decreased data throughput from teleoperated and semi-autonomous robots~\cite{murphy_disaster_2016}.

\begin{figure}[t]
    \centering
    \includegraphics[width=.85\linewidth]{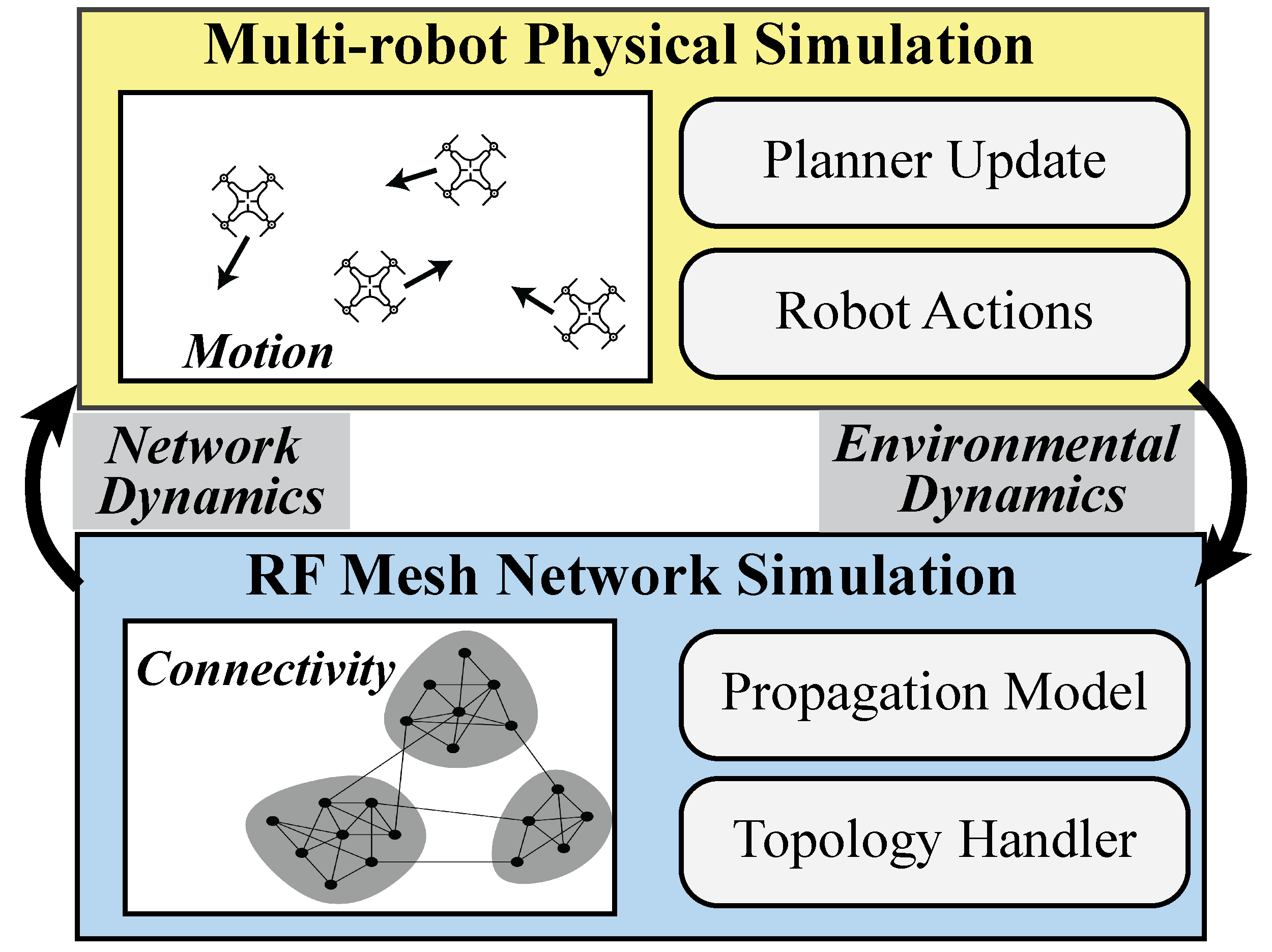}
    \caption{System diagram for \textit{\simname}, a simulator for studying the effects of accurate communications models on multi-agent control.
    The simulator exchanges information between a standards compliant networking simulator with a modular, lightweight robotics simulator.
    }
    \label{fig:blockdiag}
    \vspace{-1em}
\end{figure}

Owing to the complexities of RF communications, many robotics simulators do not include the option to evaluate control with realistic communications models. 
The most frequently used models for peer-to-peer links -- including line-of-sight, nearest neighbor communication, and perfect communication within a certain radius -- can limit effective translation to the real world.
Experiments show that many RF networks can be realistically described by extending the Friis free-space model (transmission power is inversely proportional to radius)  to an \textit{experimentally random} model for RF Received Signal Strength Indicator (RSSI) due to the complicated dynamics of a given environment~\cite{municio2019simulating,demir2019diva,tanaka2020trace}. 
Accurately determining who can communicate with whom is only one of the many parts required for translating simulated multi-agent controllers to systems with real RF communication.
Integration between the controller design and the dynamics of information routing must be considered in order to determine how precisely each agent understands the intents of its neighbors and each agent in the swarm understands progress towards the group's task.
An accurate simulation of the network is important; if the simulated network is too optimistic, the control tuning will not necessarily translate to the real world.
Conversely, if the simulation is pessimistic and the controller is tuned to accommodate a poorly performing network, it can result in sub-optimal behavior such as slow or inefficient task performance.

In this work we examine the effects that simulating accurate communications has on canonical swarm robotics tasks.
We detail the construction and use of \simname~(framework shown in Fig. 1), an extensible open source swarm robotics simulator with an embedded RF networking simulator, which is scalable to 1000s of agents with detailed logging and visualization.
Our experiments show that in decentralized formation control and flocking, varying the assumptions on RF communication (via the propagation model and scheduling functions) can make conceptually simple tasks difficult to reliably perform.

\begin{figure}[t]
    \centering
    \includegraphics[width=\linewidth]{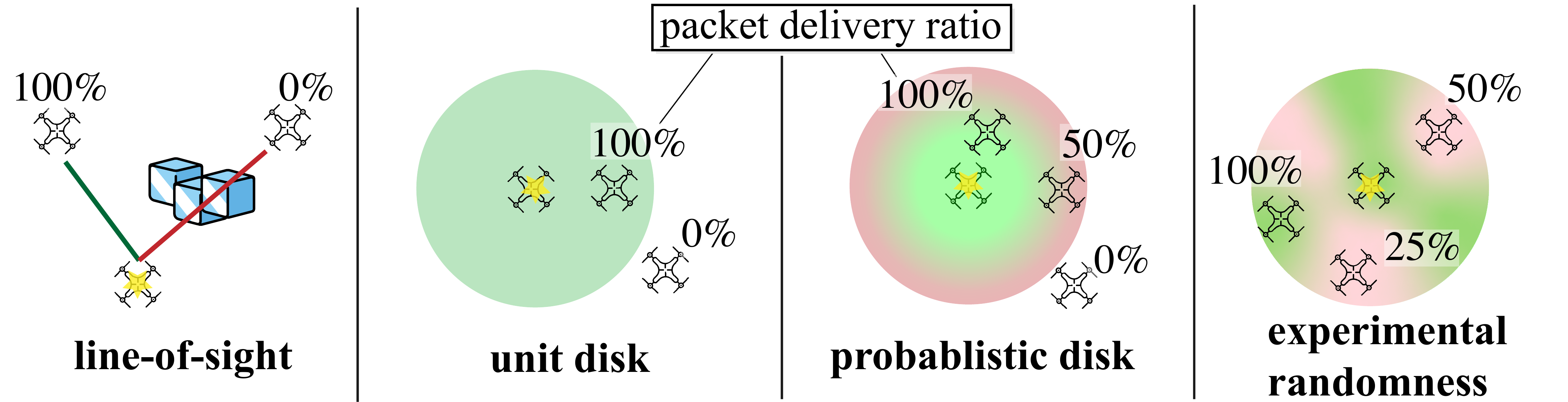}
    \caption{Different propagation models used in wireless sensor network and multi-agent control simulators.
    From left to right: 
    \textit{line-of-sight} models allows agents to connect within their field of vision; 
    \textit{unit disk} models connect with any agent within a radius $r$;
    \textit{probabilistic disk} models reduce RSSI uniformly as distance increases from communicator; and
    \textit{experimental randomness} models are derived from logging RF connections in real-world environments.
    The experimental randomness model shows that there is a much weaker correlation between position and connection than is often assumed. }
    \label{fig:prop}
\end{figure}

\begin{figure}[t]
    \centering
    \includegraphics[width=\linewidth]{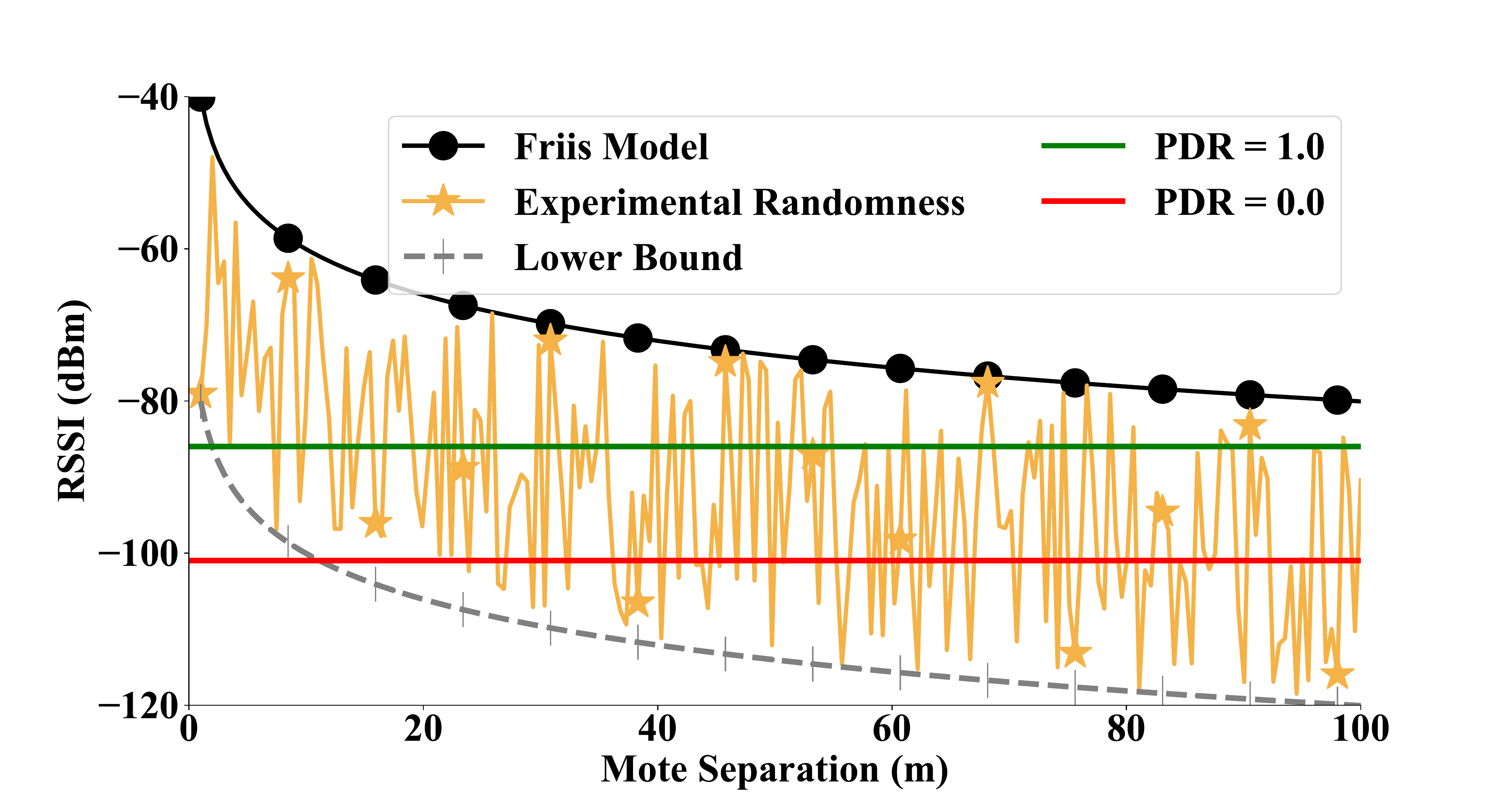}
    \caption{
    Received signal strength from a 0dBm transmitter as a function of different propagation models. 
    The Friis free space transmission loss model is an optimistic best case for unobstructed transmission. In reality, environmental factors lead to significant deviation from this ideal. 
    The \textit{experimental randomness} model, which is based on empirical data, can be used to approximate the behavior of signals in real world environments.
    The lower bound of this experimental data, approximately the Friis model prediction minus 40dBm, can be used as a conservative estimate.
    The empirical model, however, shows that the true signal strength will often be better than this, providing opportunity for more optimistic control patterns\protect\footnotemark.
    }
    \label{fig:pisterhack}
\end{figure}

\section{Background}
\label{sec:background}
\label{sec:wsn}
In this section we detail the background needed for understanding the interplay of network dynamics with multi-agent systems.

Wireless sensor networks (WSNs) are connections between many individual radios, referred to as \textit{motes} or \textit{nodes}.
Two motes have a Packet Delivery Ratio (PDR), denoting the number of received packets out of the total sent, which is often heavily influenced by the Received Signal Strength Indicator (RSSI). 
In simulated networks, the relationship between physical distance and RSSI or PDR is called a \textit{propagation model}.
The algorithm controlling which mote is speaking at a given time-step is called a \textit{schedule function}.
For a more detailed review of sensor network behavior, see~\cite{municio2019simulating}.

\footnotetext{The connectivity traces were obtained from the \url{http://wsn.berkeley.edu/connectivity/ project} at the University of California, Berkeley. 
The data set used is “soda,” created by Jorge Ortiz and Prof. David Culler.}

\subsection{RF Propagation Models}
Determining the RSSI between network nodes is crucial to accurately simulating communication, as it is the primary driver of PDR.
Using theoretical abstractions for predicted signal strength in simulators is necessary because RF signal propagation is extremely difficult to model.
Different propagation models are visualized in \fig{fig:prop}: 
line-of-sight sets PDR to be 1 for any agents with an unobstructed visual link, 
a unit disk delivers all packets within a radius $r$, and 
a probabilistic disk creates a lower bound on communication radius by matching the lower bound of free path loss.
Historically, the RSSI, which underpins propagation models, is  very hard to model due to the complexities of electromagnetic propagation, but recent work has shown the accuracy of an \textit{experimental randomness} model~\cite{municio2019simulating,demir2019diva,tanaka2020trace}, shown in \fig{fig:pisterhack}. 
In this work we show the effects of different unit disk radii, but focus on a radius of $r=10\text{m}$ because that is near to where the PDR hits 0 for the lower bound in \fig{fig:pisterhack}. 
Changing communications hardware (e.g., to higher power transmitters or more sensitive receivers) would correlate to a vertical shift of the RSSI at a given distance.

\subsection{6TiSCH Networks}
In this work we focus on 6TiSCH, a standards-compliant RF mesh networking protocol which is designed to be low-power, demand minimal computation overhead, and able to integrate with existing internet services. 
Together, these features make it well-suited for use in high agent-count, low-cost autonomous systems.
6TiSCH combines Time Synchronized Channel Hopping (TSCH) with Internet Protocol version 6 (IPv6)~\cite{municio2019simulating}.
Time Synchronized Channel Hopping (TSCH) changes when and which channels are used in wireless communications to provide more reliable communications.
A group of channel options combine to form a slotframe, where a slot is a fixed time width interval at a specified channel frequency.

\begin{figure}[t]
    \centering
    \includegraphics[width=\linewidth]{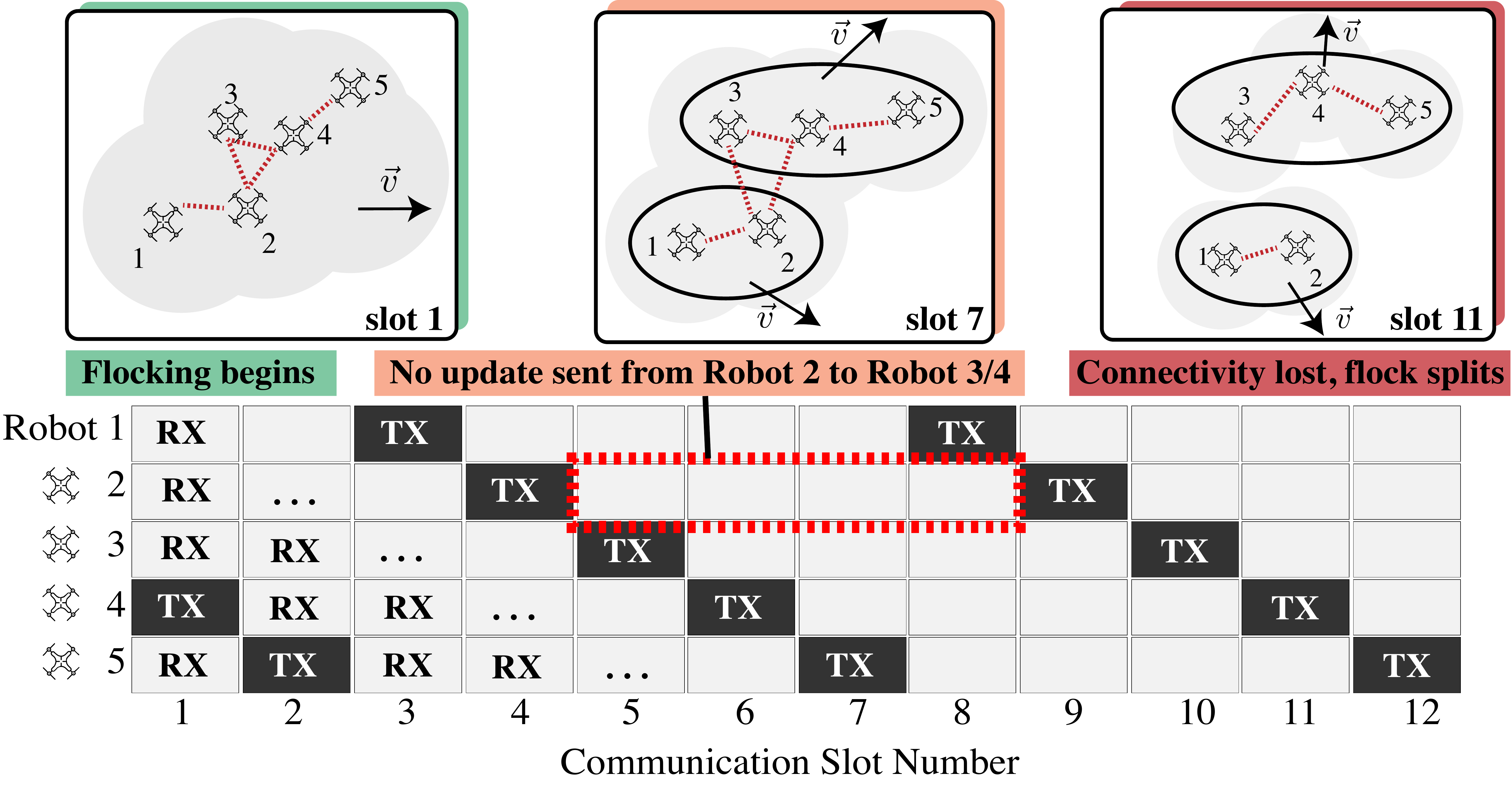}
    \caption{Suboptimal schedule functions rapidly deteriorate swarm performance.
    Below is a round robin scheduling function (RRSF) grid for per-agent transmit or receive behavior, where agents broadcast position to all agents within range.
    Due to the asynchronicity of networked communication, a delay in broadcasting can result in the position of the important agents not being updated in a crucial timewindow.
    }
    \label{fig:schedule}
\end{figure}

\subsection{Network Scheduling Functions}
The control mechanism for who transmits (\textit{TX}) and receives (\textit{RX}) during each slot in a slotframe is called the schedule function.
This schedule is necessary because multiple motes transmitting on the same channel, or a mote designated to transmit and receive simultaneously, regularly results in dropped packets.
When starting a task, the scheduling function determines the initial communications network structure as peers negotiate who will speak when.
Different scheduling functions have substantial variance in the amounts of time to form and the resulting connection structures.
How best to form a network is an open avenue of research investigation beyond the scope of this paper, and such structure changes can be very important for the downstream task.

The 6TiSCH network protocol was designed with a Minimal Scheduling Function (MSF) to negotiate connections over time with stationary nodes.
Our experiments show limitations with using MSF for communication between mobile nodes, namely because MSF is designed to slowly converge to a slotframe schedule for static nodes based on negotiations, transmission collisions, and PDR measurements over several slotframes. 
For mobile nodes using this scheduling function, communication links will frequently break down due to rapid position changes of agents. 
This issue becomes further exacerbated when scaling to larger numbers of agents because there are only a finite number of slots where any given agent can update its neighbors with its current position data.

To improve on network stability with the MSF, we have implemented a round robin scheduler function (RRSF), shown in the bottom of \fig{fig:schedule}, where every new slot schedules one mote to transmit while all other motes are in receive mode, cycling through each mote over the course of a slotframe. 
Compared to this new RRSF, the MSF is slower to converge to an initial network topology because it entails peer-to-peer negotiation over transmit-receive slots, and re-negotiations make it substantially less stable to the changes in network architecture which arise with mobile WSNs.
Even with RRSF, challenges in multi-agent control can arise when crucial agents do not send a packet, resulting in link loss; this challenge of communications in swarm control is shown in \fig{fig:schedule}. 

\section{Related Works}
\label{sec:related}

\subsection{Mobility in Wireless Networks}
A typical wireless sensor network (WSN) is designed and optimized as a stationary system. 
In networks composed of mobile nodes, many events can result directly from physical movement, including connection topology changes, removing or adding of network members, unexpected node failures, signal strength change, and more~\cite{mehta2012mobility,silva2014mobility}.
Methods to improve sensor network performance in the presence of mobile nodes come at many layers of the communication stack, but primarily can be characterized at the Medium Access Control (MAC) layer -- who communicates when -- and at the network layer when routing -- determining how to distribute information over multiple steps through the network.
Solutions at both of these layers are highly application-specific and difficult to design and apply to the general case.
For example, tools at the MAC layer optimize for mobility by being selective about which packets are sent, for example by using tools for avoiding packet collisions~\cite{ali2005mmac,khan2013collision}.
Scheduling functions are used to control for transmission timings and are also crucial for maintaining a network during movement, so they can also be tuned for mobility, such as efficient scheduling with delay constraints~\cite{gu2012eswc}.
Multi-hop routing for mobile WSNs is an open area of research, with solutions investigating moving nodes to pass information to another area of the network~\cite{vincze2005multi}, but little work has been performed optimizing specifically for multi-robot control tasks.

Network node mobility can also be used in a complimentary optimization problem, for example by incorporating network maintenance and health directly into robotic task planning.
Route Swarm~\cite{route-swarm} adds a second optimization to a coordination problem in order to maximize information flow across a network. Mobile nodes can also be used to ferry messages between stationary nodes~\cite{robot-msg-ferrying}.

\subsection{Communication in Multi-agent Robotics}
Groups of robots can work together and accomplish tasks unrealistic for a single unit, often by leveraging the information shared by their neighbors~\cite{balch1994communication}.
The potential of peer-to-peer communication is highlighted by progress in decentralized control, but recent work has incorporated only a limited study of realistic propagation models, schedule functions, multi-hop behavior, and other real-world aspects of complex networks.
The simplest network abstractions used in robotics studies focus on swarms as a graph structure, using a model where the nearest neighbors can communicate fully~\cite{moreau2005stability,zhang2018fully}.
Some work uses line-of-sight (LOS) instead of purely geometric distance in order to determine which agents can communicate~\cite{arkin2002line,zelazo2012rigidity}, but LOS is not an accurate model for RF networks.
The most common approach to communications may be a radius of connectivity model, where all neighbors within a radius can communicate~\cite{de2006decentralized,yang2008multi,su2009flocking,filippidis2012decentralized}, but this assumption uses inconsistent communication radii rather than information inspired by real systems. 
More advanced communications models used in the literature include a Gaussian well model, where RSSI and PDR decrease with increasing radius, but examples incorporating these remain rare~\cite{jimenez2018decentralized}.
\simname~provides a tool that allows for simple and modular manipulation of the propagation models to show how the aforementioned communications assumptions change behavior of multi-agent tasks. 
The different propagation models common in the robotics literature and studied with our framework are shown in~\fig{fig:prop}.

\subsection{Simulating Networked Swarm Robotics}
Accurately simulating all aspects of a robotic swarm, including the network, robots, and environment, is a challenging task.
Many tools for multi-agent simulation specialize at one function -- such as Gazebo~\cite{Koenig-2004-394} or AirSim~\cite{airsim2017fsr} for accurate robot dynamics, SwarmSim~\cite{9199255}, SwarmLab~\cite{soria2020swarmlab}, or Scrimmage~\cite{demarco2018} for swarm control, and the OpenWSN emulator~\cite{watteyne2012openwsn} or the 6TiSCH Simulator~\cite{9199255} for realistic communications models.
There has been relatively little development of simulation tools at the intersection of robotic control and accurate network modelling.
RoboNetSim~\cite{robonetsim} and ROS-NetSim~\cite{rosnetsim} are two frameworks designed to mediate external simulators and synchronize data for accurate communications in robotics.
RoboNetSim acts as an integration tool between any two robotics and networking simulators. 
For example, RoboNetSim synchronizes time and robot position between Network Simulator 2 (NS-2)~\cite{issariyakul2009introduction} and ARGoS~\cite{Pinciroli:SI2012}, a popular multi-agent simulator.
RoboNetSim is no longer actively supported.
ROS-NetSim performs a similar function to RoboNetSim by creating a central Robot Operating System (ROS~\cite{quigley2009ros}) node between the network and robotic simulators. 
ROS-NetSim has promise for improving sim-to-real transfer because many real systems are deployed on ROS~\cite{quigley2009ros}, but is not designed to handle high agent counts nor standards compliant communication models, which is helpful for transitioning to real WSNs.
Our simulator, \simname, allows rapid experimentation and modularization by having built-in network and robotic simulation, is written entirely in Python for ease of development, and is scalable to 1000s of agents.

\begin{figure}[t]
    \centering
    \begin{subfigure}[t]{\linewidth}
        \centering
        \includegraphics[width=\linewidth]{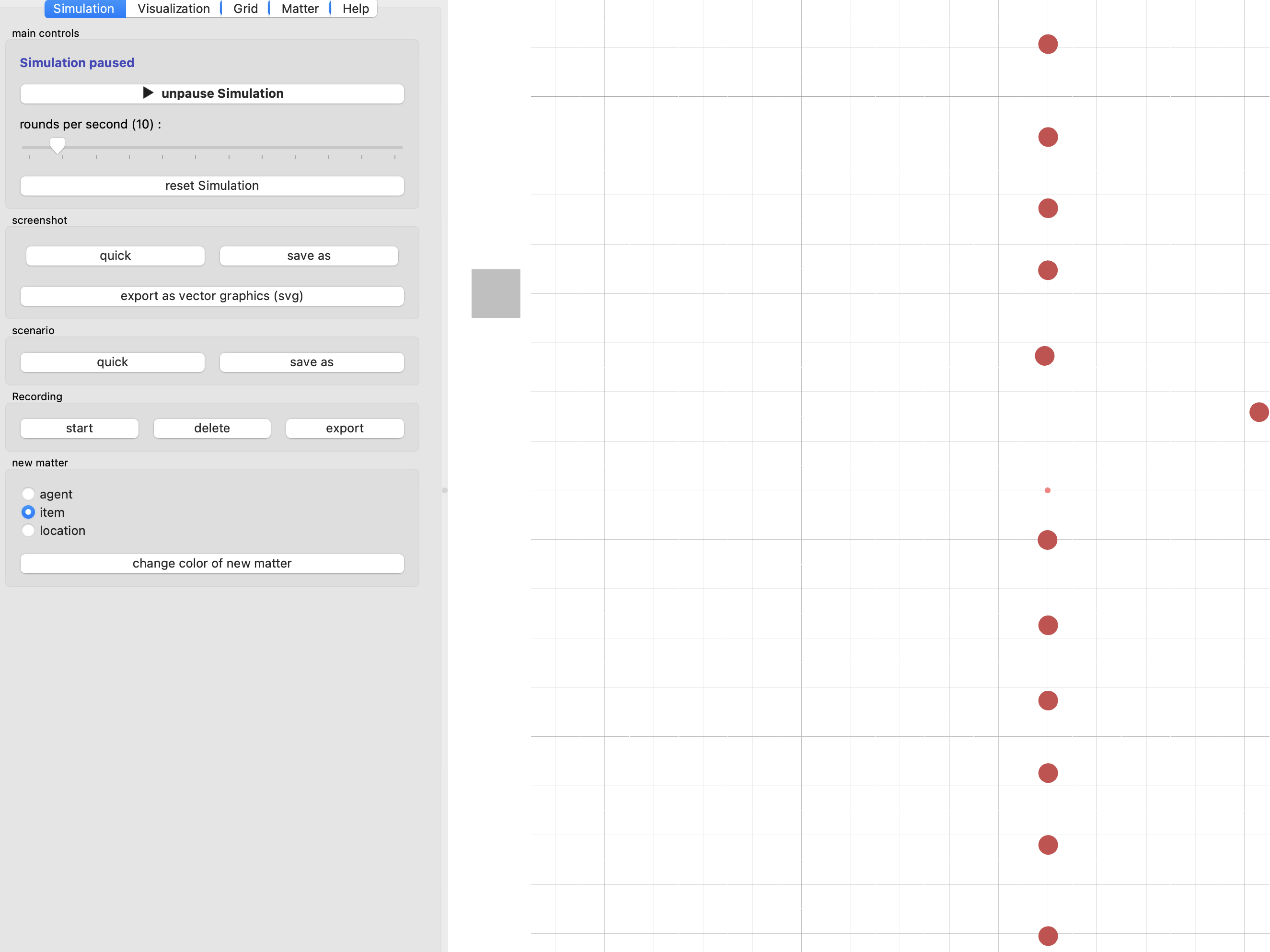}
        \caption{Robot dynamics visualizer.}    
        \label{fig:swarm-viz}
    \end{subfigure}
    \\
    \begin{subfigure}[t]{\linewidth}  
        \centering 
        \includegraphics[width=\linewidth]{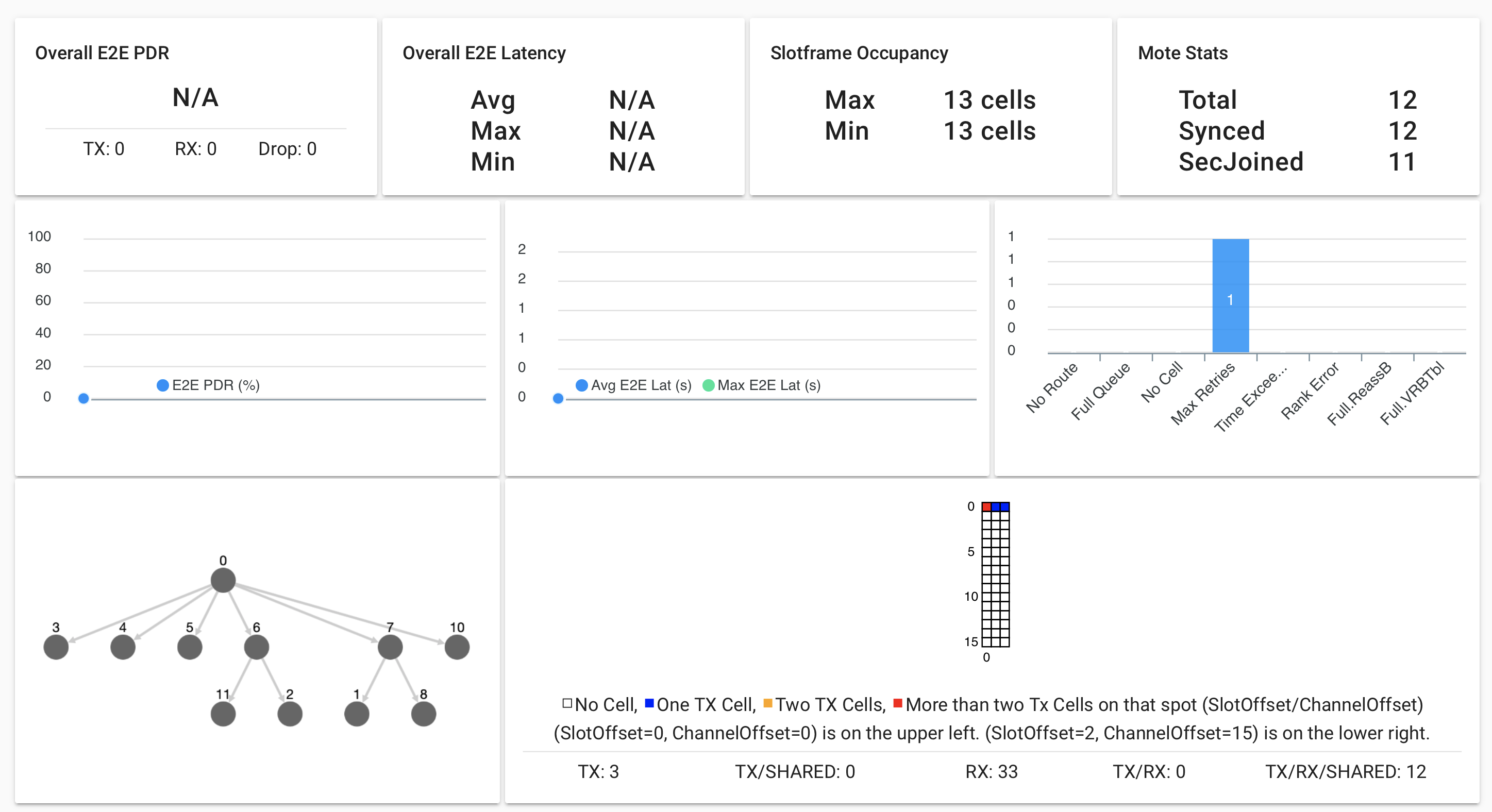}
        \caption{6TiSCH network visualizer.}    
        \label{fig:network-viz}
    \end{subfigure}
    \caption{\simname~is configurable to visualize continuous network and/or robotic dynamics.
    A core component of the network visualization, the connection graph, is shown in the bottom left.
    Lopsided network configurations can make a swarm sensitive to dropping members from the network during complicated multi-agent tasks.
    For more details on the robotic simulation, see~\cite{9199255}, and for details on simulating 6TiSCH networks, see~\cite{municio2019simulating}.
    }
    \label{fig:viz}
\end{figure}

\section{\simname: Swarms with 6TiSCH Networks}
\label{sec:simulator}
In this section we describe the open source simulator we have released, \simname, which is suitable for studying large numbers of mobile networked whileagents with realistic communications models. 
The simulator, visualized in \fig{fig:blockdiag}, uses the environment and agent abstractions from SwarmSim~\cite{9199255} and simultaneously integrates a IEEE802.15.4 6TiSCH simulator~\cite{municio2019simulating}.
In building the simulation framework, we define an abstract class through which 6TiSCH can interface with a multi-agent robotic simulator using remote procedure calling (RPC). 
This allows for the transfer of the minimal set of necessary data to guarantee time and state synchronization between simulators. 
It can be extended to transfer any type of useful information between the network and physical layer simulators, for example allowing for robotic and network routing control algorithms to depend on one another,
network failure information being exchanged to take appropriate actions at the robotic controller level, and for 
physical environment information to be communicated to the network simulator to inform propagation modelling. 
The code for the simulator and experiments can be found at \code.

\begin{table*}[t]
\centering
\begin{tabular}{  p{1cm} | p{3cm} p{4cm} p{4.2cm} }
\toprule 
 Agent \newline Count & RRSF Network Formation \newline (Simulation Timesteps)  & Full Network Simulation Speed \newline (Clock Time Per-step, ms) & Propagation Model Simulation Speed \newline (Clock Time Per-step, ms)  \\ 
 \midrule
5 & $141.4\pm21.8$   &  \num{6.255} & \num{0.85} \\
10 & $187.8\pm71.4$   &  \num{21.45} & \num{1.6} \\
25 & $418.8\pm488.2$   &  \num{127.2} & \num{4.1} \\
50 &  $2485.4\pm2221.4$  &  \num{495.5} & \num{8.3} \\
100 &    &  & \num{17.9} \\
250 &    &  & \num{55.2} \\
500 &    &  & \num{153.8} \\
1000 &   &  & \num{560.3} \\
2500 &   & & \num{2273} \\
  \bottomrule
 \end{tabular}
 \caption{\simname~performance characteristics. 
  The mean and standard deviation of steps reported for network formation are shown to highlight the variance of network formation with a set scheduling function.
  10 trials per configuration were executed with agents initialized in a line with \SI{2}{\meter} spacing.
   Due to the limitations of schedule functions with a full network simulator, high agent count ( >100) networks failed to converge with the full network simulator and the RRSF regardless of propagation model.
 Columns two and three indicate the mean period per dynamics step when running different numbers of agents (the standard deviations of clock-times per step are extremely low and omitted).
 The Full Network Simulation Speed is when using the full 6TiSCH simulator, but this does not scale well to high agent counts due to the challenges of scheduling function design. 
 To accommodate high agent counts, the simulator can instead run with a limited communications module that can vary the propagation function (``Propagation Model Simulation Speed''), which is where experiments can scale to 1000s of agents.
 These experiments are run with a 2.3 GHz 8-Core 9th-generation Intel Core i9.
 }
\label{tab:performance}
\end{table*}

\subsection{Design}

\textbf{Building on SwarmSim}:
The dynamics simulation for \simname~extends SwarmSim's minimal infrastructure~\cite{9199255}.
To capture more realistic dynamics, we created an agent structure \texttt{VeloAgent} that has continuous velocity-controlled dynamics instead of the SwarmSim default with only positional dynamics.
This extension can be abstracted further to accommodate more realistic robotic simulations.
We modified the \texttt{World} class to pass arguments into the simulator environment for configurable experimentation. 
Additionally, we fixed multiple bugs preventing out-of-the-box examples in SwarmSim from running.

\textbf{World Dynamics}: 
\simname~has a world consisting of two portions: network and environment dynamics. 
The network dynamics dictate which agents can communicate and what information is passed between them. For example, with the experimental randomness propagation model, when the robot simulator moves an agent in space, the propagation model updates its RSSI distribution per each mote-to-mote link as a function of separation to determine the network dynamics.
The communications dynamics can be expanded to study how data flows through multi-hop networks.
The environment dynamics controls how each agent moves through space, and adds disturbances or constraints on motion.
These can be modified by further instantiating the class \texttt{Scenario} (inherited from SwarmSim), such as for when the robotic task is heavily dependant on the environment (e.g., indoor space exploration). 
Finally, additional varieties of robot dynamics can be encoded by adding subclasses of \texttt{Agent}.

\textbf{Controllers}:
We expand the original SwarmSim architecture of a controller class called \texttt{Solution}~\cite{9199255}. Agents can be controlled using both the global environmental information and via their internal belief states, for example of who their nearest neighbors are based on networking data. Alternatively, both the networking simulator and the environmental simulator can send controls directly to agents through their RPC endpoints in order to study event-driven control.

\textbf{Communications}: 
The communications stack is designed to study both low-level RF dynamics (who can communicate with whom) and multi-hop network behavior (how to route packets through the swarm). 
This paper focuses on the peer-to-peer communication problem, dictating to each agent which neighbors it has successfully transferred data with in a given slotframe, but the 6TiSCH network simulator is designed to be able to study multi-hop packet routing.
The low-level encoding of bits into packets and maintaining IPv6 standards are handled entirely within the 6TiSCH simulator.
The communications stack can interface with controller design in multiple ways.
For example, included in~\simname~are different options for when controls are updated with respect to communication -- including after each packet, after each slotframe, or at a user-specified fixed slot-frequency.
For high agent-counts, network initialization is intractable with simple scheduling functions (e.g., due to frequency of packet collisions), so experiments can also be run only evaluating the difference between propagation models.

\subsection{Usage}
\textbf{Experimentation}:
The publicly available \simname~repository includes many environment scenarios and control solutions included at launch to foster use in multiple application domains of multi-agent systems.
The included scenarios and solutions for flocking and formation control are easily used with a different agent, world, or controller: all environments and controllers are modular and changed via a configuration file. 
A quickstart guide and instructions for running batched experiments are included in the repository.

\textbf{Visualization}:
The 6TiSCH network visualizer runs in the browser and dynamically shows which agents are connected, the schedule matrix, packet delivery over time, and more~\cite{municio2019simulating}.
SwarmSim includes a OpenGL based video rendering tool to show the physical dynamics in 2- or 3-dimensions~\cite{9199255}.
With the RPC server, we can synchronize and visualize these two diagnostic tools simultaneously, as shown in~\fig{fig:viz}.
Currently, only the dynamics visualization can be saved and exported as a video, but all network and dynamic data are saved individually after each experiment as compressed arrays in \texttt{.dat} files (the frequency of saving and file type is configurable).
Future work will address simplifying and merging this logging system.

\textbf{Scaling \& Performance}:
\simname~is capable of scaling to 100s of agents for rapid iteration on experiments with different control tasks, communication models, and robots.
The simulator update period per dynamics step is shown for different numbers of agents in \tab{tab:performance}.
To show the challenge of network formation, we have shown the number of timesteps it takes for the network to converge and indicate which configurations do not converge with the 6TiSCH simulator.
For experiments where the network does not converge within a reasonable time, we have included the per-agent simulator scaling when running the dynamics engine with only propagation models (assuming all agents can send and receive packets simultaneously).
Performance scaling per-step degrades dramatically when using the real-time visualization tool.

\begin{figure}[t]
    \centering
    \includegraphics[width=.95\linewidth]{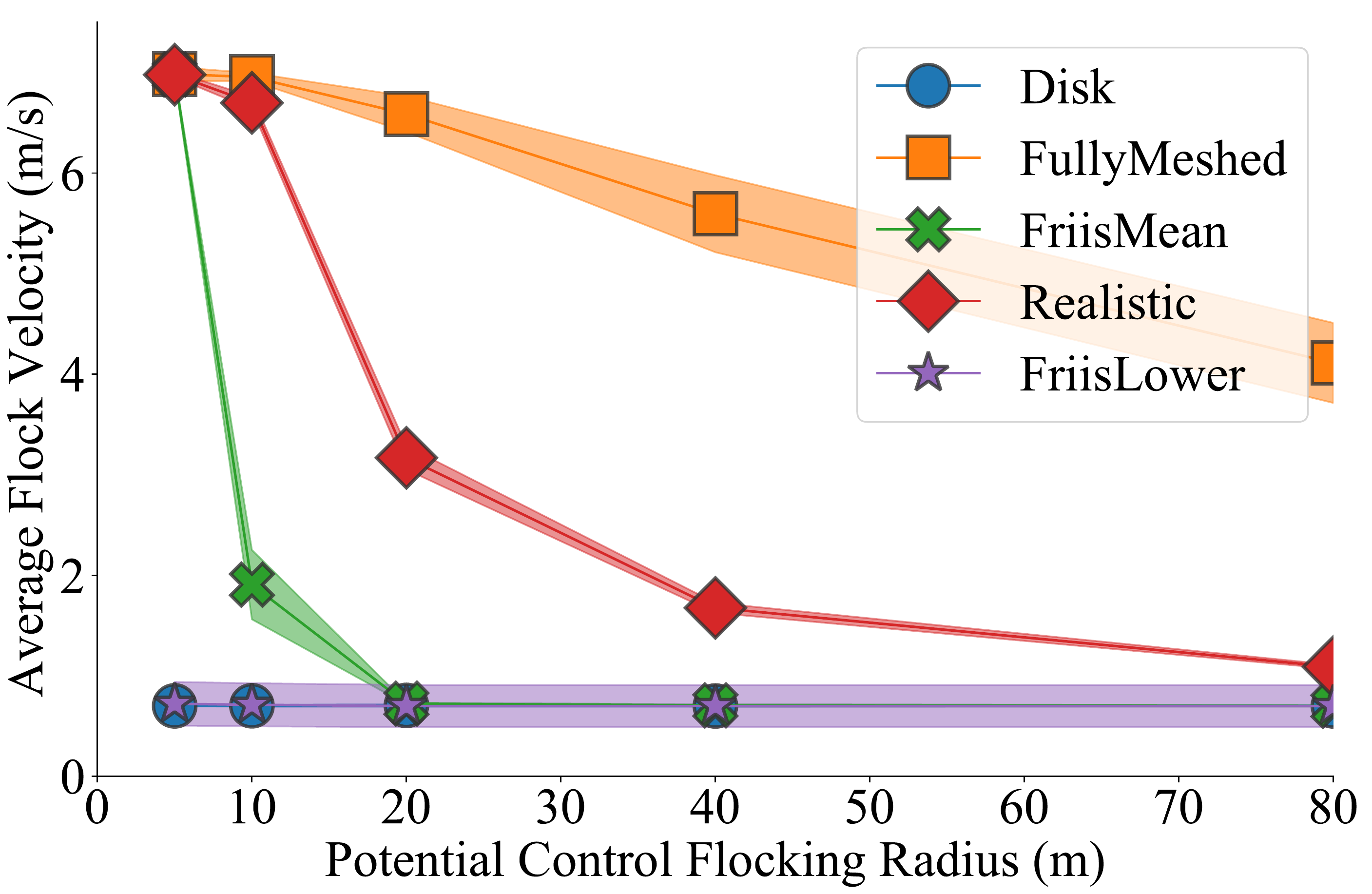}
    \caption{
    Median and standard deviation flocking velocity for different potential control connectivity radii (10 trials per configuration, 10 agents per trial, RRSF). 
    Average velocities close to zero generally indicate that the flock fails to follow the leader agent. 
    The results show that using more realistic propagation models in the networking simulation stack sets bounds on the potential control flocking radius; at higher radii, agents lose connection from the leader and potentially their neighbors (depending on collision radius).
    }
    \label{fig:flock_radius}
\end{figure}

\section{Results}
\label{sec:exp}
We have quantified and compared the performance of multi-agent flocking and formation control over a set of different communication paradigms (full connectivity, unit and probabilistic disks, and the experimental randomness model).
The agents used for our experiments have simple point-mass dynamics with a maximum velocity of \SI{30}{\meter \per \second}.
All experiments use a round robin scheduling function (RRSF). 
The reported performances are averaged over the timesteps after networks are formed, which varies with different scheduling functions, propagation models, and control tasks.

\begin{figure}[t]
    \centering
    \includegraphics[width=.95\linewidth]{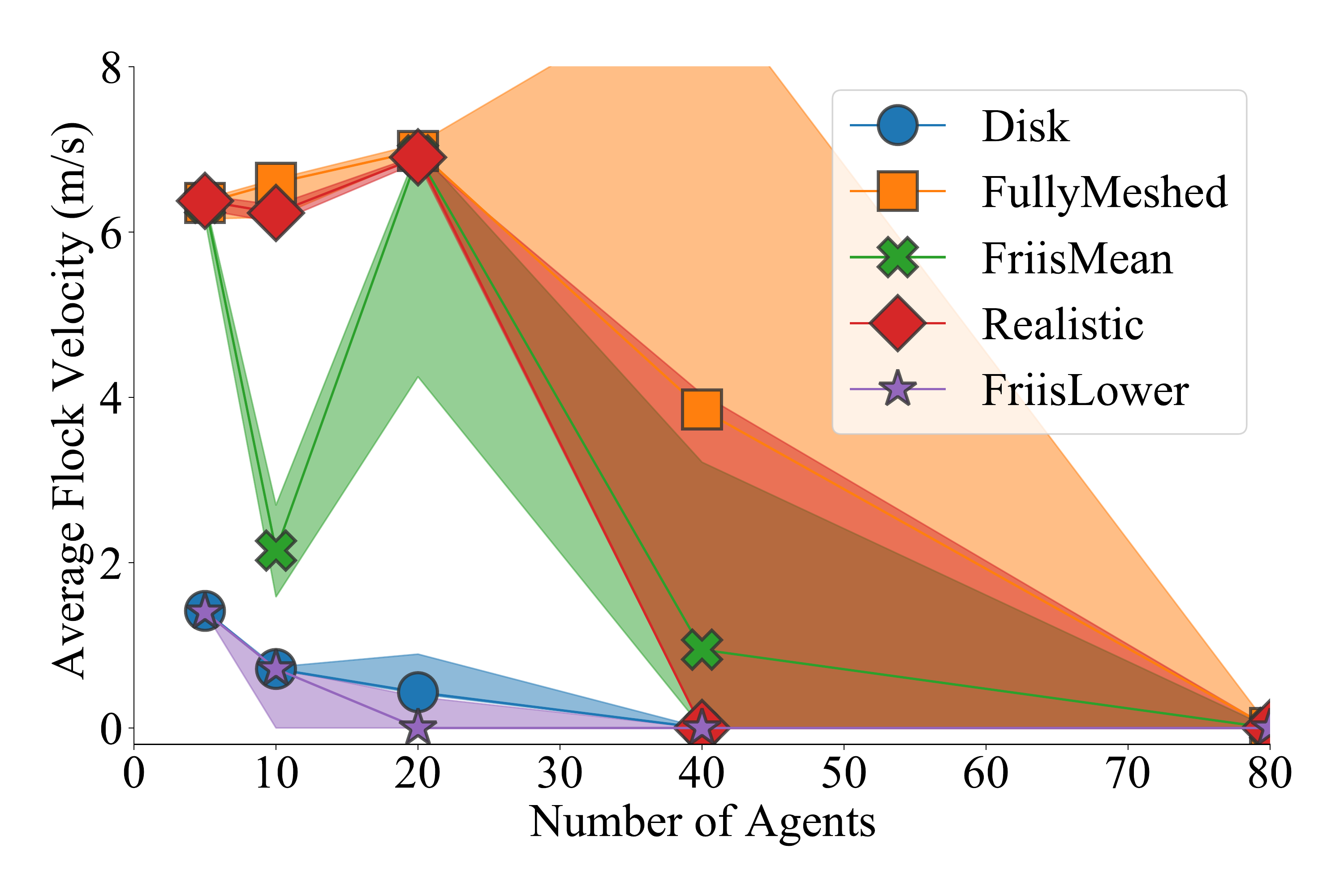}
    \caption{
    Flock speed (median with maximum and minimum across trials) versus number of agents with ideal vs real communications models (10 trials per configuration, $r_{flock}=10\text{m}$, RRSF).
    There is a complicated relationship between number of agents and agent separation that results in a large variance in potential outcomes of flocking.
    A minimum of 0 is a trial where the agents failed to form a stable network, prohibiting flocking from starting.
    }
    \label{fig:flock1}
\end{figure}

\begin{figure*}[t]
    \centering
    \begin{subfigure}[t]{0.27\linewidth}
        \centering
        \includegraphics[width=\linewidth]{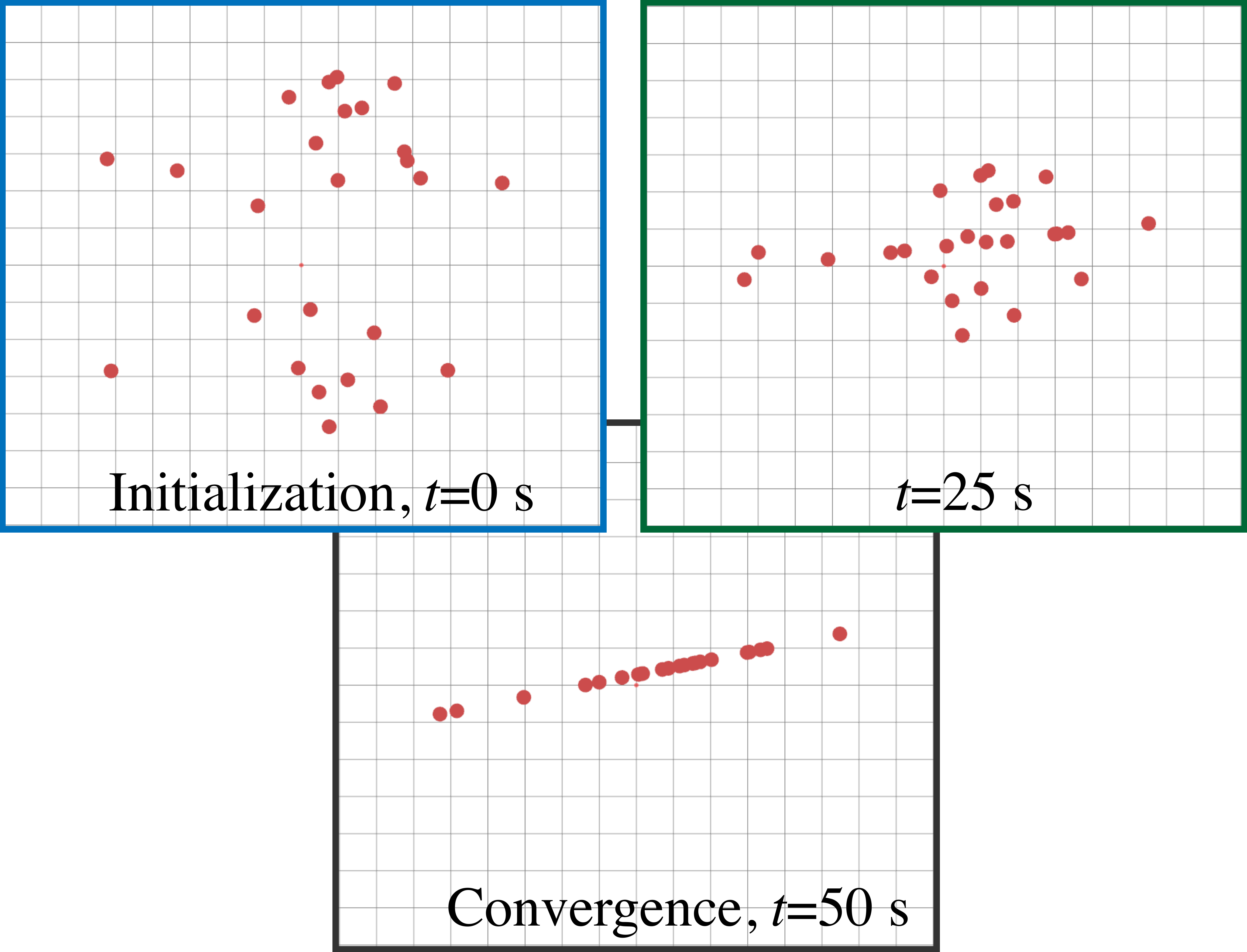}
        \caption{Example with 25 agents. 
        }
        \label{fig:form-ex}
    \end{subfigure}
    \hfill
    \begin{subfigure}[t]{0.34\linewidth}
        \centering
        \includegraphics[width=\linewidth]{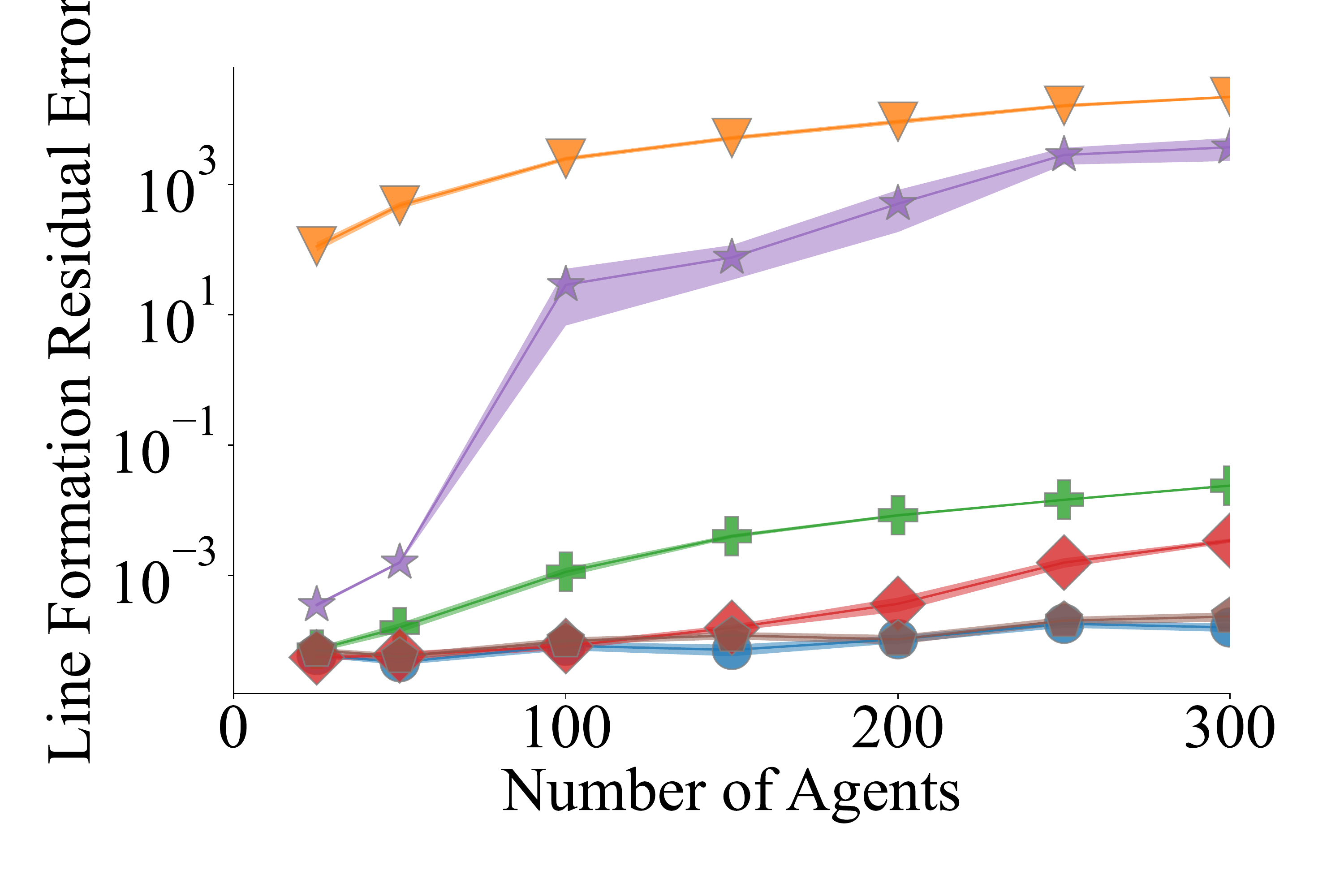}
        \caption{Log-scale error from the desired formation. 
        }
        \label{fig:form-error}
    \end{subfigure}
    \hfill
    \begin{subfigure}[t]{0.34\linewidth}
        \centering
        \includegraphics[width=\linewidth]{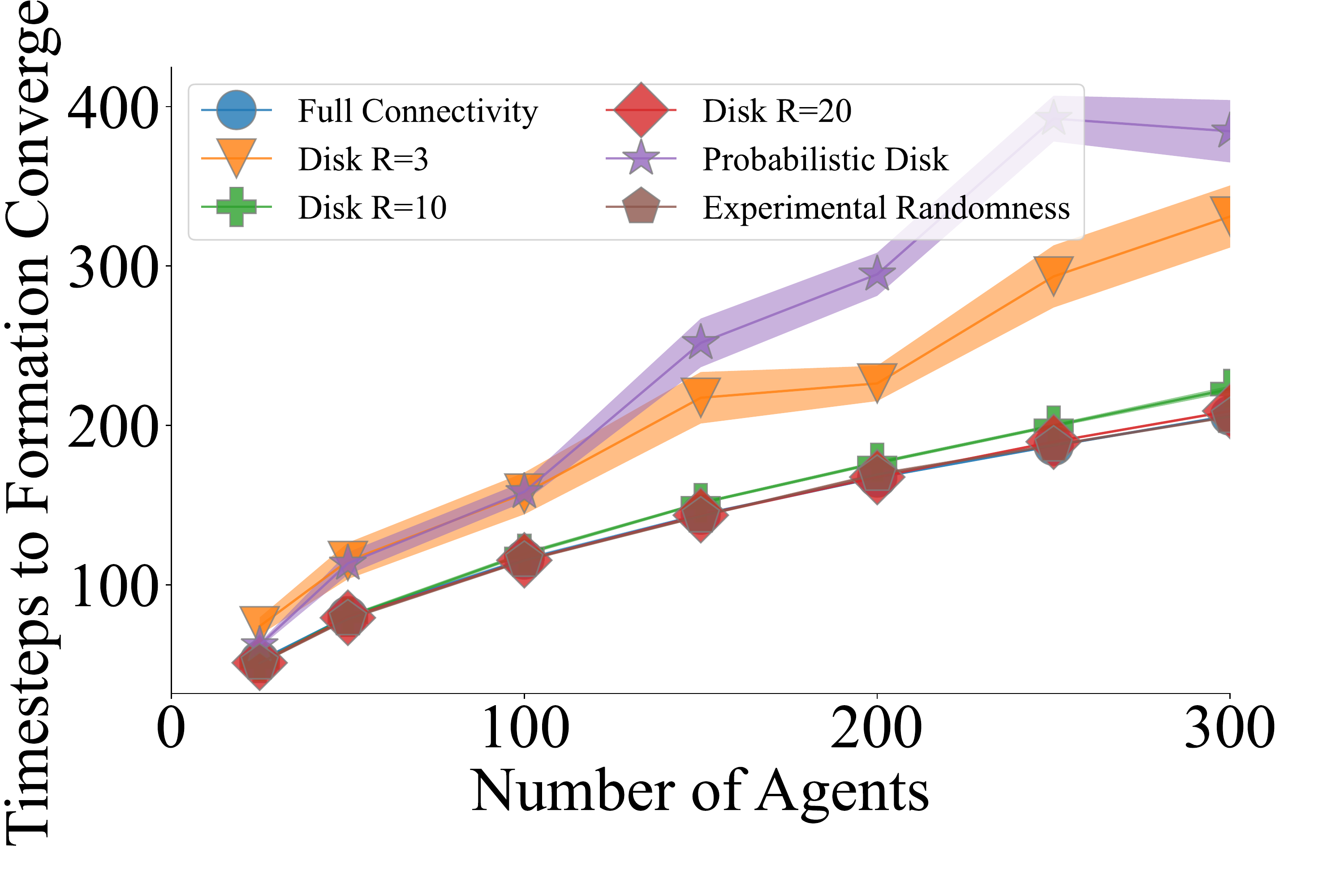}
        \caption{
        Time until convergence.
        }
        \label{fig:form-time}
    \end{subfigure}
    \caption{
    Example visualization and quantitative performance of decentralized line formation control with different communications models and number of agents (16 trials per configuration and using propagation model only to avoid the intractability high-agent count network formation; median and standard deviation are shown). 
    (a) three time-snapshots from an example run of the line-formation control task with \simname's built in visualizer.
    (b) the magnitude of error with different communications models during line formation. 
    This error arises when sub-groups of agents form multiple lines due to dropped communication with their neighbors. 
    (c) the time to completion of the task, \textit{i.e.} when no agent is off their projected target at a given time, varies more closely with the number of agents.
    Full connectivity in this example represents the minimum time to completion and error for the randomly initialized agents, and the experimental randomness model maintains performance at high agent counts.
    }
    \label{fig:form}
\end{figure*}

\subsection{Flocking}
\label{sec:flocking}
Flocking is a common task in decentralized multi-agent control where members of the group must co-align with their neighbors and therefore maximize the average velocity in a desired direction. 
In this section, we show how the task can fail, either via the flock failing to form or moving with very low velocities, due to imperfect communications and the resulting delayed or erroneous control updates.
Given a potential function, $V_{ij}$, valuing a set distance between any two agents, flocking can be formulated as an optimization problem over the controls of each agent $u_i$, given the velocity $\vec{v}_i$ and position $\vec{x}_i$ of each agent~\cite{zavlanos2007flocking}:

\begin{align}
    u_i &= - \sum_{j\neq i} (\vec{v}_i-\vec{v}_j) - \sum_{j\neq i} \nabla_{\vec{x}_i} V_{ij} 
    \label{eq:flock} \\
    V_{ij} &= \frac{1}{\|\vec{x}_i-\vec{x}_j\|_2^2} + \frac{1}{r_\text{flock}^2-\|\vec{x}_i-\vec{x}_j\|_2^2}
    \label{eq:flock-v-orig}
\end{align}
The first term is for directional alignment and the second term minimizes collisions (outward pressure) and maintains network connectivity (inward pressure).
In our simulations with robots capable of high velocities, the discrete control update often reached a singularity at the flocking radius in \eq{eq:flock-v-orig} when agent separation increased beyond the desired $r_\text{flock}^2$ before a position update, leading to flock divergence (even if connected).
For this paper, we use a modified, singularity-free flocking value function $V_{ij}$ to enable flocking with discretized control inputs with varying network connectivity:
\begin{align}
    \widetilde{V}_{ij} & = \sum_{j}K_\text{col.}e^{-\frac{\|\vec{x}_i-\vec{x}_j \|^2_2}{r_\text{collision}^2}} + K_\text{conn.} e^{\frac{\|\vec{x}_i-\vec{x}_j \|^2_2}{r_\text{flock}^2}},
    \label{eq:flock_grad}
\end{align}
where $K_\text{col.} = r_\text{collision}^2+r_\text{flock}$ and $K_\text{conn.} = r_\text{flock}$.
Two characteristic radii are important to flocking behavior: the collision radius between agents, $r_\text{collision}=0.8\text{m}$ for all experiments, and the maximum desired separation between connected members, $r_\text{flock}$, which we vary in experiments.
Even with the stabilized flocking gradient in \eq{eq:flock_grad}, emergent flocking failed to converge regularly when not using the full connectivity propagation model.
To create a more stable flocking task, we set the initial magnitude and direction of one \textit{master} agent, $\vec{v}^*$, and direct the remainder of the flock to maintain connectivity with the following velocity-based control inputs:
\begin{align}
    u_k & = \vec{v}^*, \quad u_i  = - \sum_{j\neq i} \nabla_{\vec{x}_i} V_{ij}.
\end{align}
When using over 10 agents, in order to maintain the flock, the weighting of the leader's velocity in the follower's flocking summation is increased by $\frac{n_\text{agents}}{10}$.
With this new follow-the-leader flocking, where we are testing a network's ability to dynamically maintain connectivity, we examine the mean flocking velocity versus the desired flocking radius and connectivity models.
A higher flocking radius leaves the agents less likely to collide, but makes agents more likely to lose connection.
The behavior of agents and stability of the behavior is sensitive to the relative magnitudes of these radii.

The flock speeds with different maximum flocking radii (\textit{i.e.} the separation that agents are pressured to stay below) are shown for the studied propagation models in \fig{fig:flock_radius}.
A similar experiment varying the number of agents is shown in \fig{fig:flock1}, where the realistic propagation model again improves on a disk model, though the relationship to flock velocity is noisy and warrants further investigation.
Mean velocity shows the capability of all agents to follow the identified leading agent, but it does not explicitly capture some emergent flocking behavior such as oscillating agents towards and away from the current swarm centroid (these mechanism do slow down the mean velocity by directing movement of the agents to be less co-aligned with the leader).
The experimental randomness model represents a substantial gain over unit disk models designed for a conservative representation of communication abilities.
With the MSF, properly quantifying the performance of flocking is difficult because the network often failed to register an agent's nearest neighbors, removing the separating force of the potential function and collapsing the flock. 
While this would result in a higher average velocity, it represents a divergence from desired behavior.

\subsection{Formation Control}
A common area of study in multi-agent systems is that of forming and maintaining pre-defined shapes through local communication~\cite{oh2015survey}. 
The experiments in this section only use the propagation model to allow scaling to higher agent counts (recall the challenge of network formation with simple propagation models discussed in \sect{sec:simulator}).
When communication is limited, the task can take more time to converge or even fail, as without a global critic, local agents do not know the success of their controller.
In this paper, we deploy agents that seek to organize themselves into a single line, irrespective of any particular ordering of agents or orientation of the line itself. 
The decentralized line formation algorithm follows from~\cite{yun1997line}: 
at each timestep, every agent locally applies least-squares to its own position and that of all neighbors it is connected to; this produces a local approximation of the optimal global line of convergence for each agent, which it uses to update its next control input.
An example of the task is shown in \fig{fig:form-ex}.
When agents drop from communication, consensus among desired planes of convergence breaks down, and sometimes leads to multiple sub-formations being formed.

We benchmark the effects of communication modeling on this task through two metrics: convergence time and residual error. 
Convergence time is defined as the number of timesteps taken until all agents stop moving (i.e., until all agents individually believe the line formation task has been achieved). 
Note that this captures only the local belief state of the agents, and not the true global quality or convergence of the agent formation into a line. 
Residual error, calculated directly from applying a standard line-fitting least square procedure over the positions of all agents, shows the relative accuracy of a given formation. 
During initialization, we randomly spawn a given number of agents within an adaptive circular region centered around the origin such that the density of agents is approximately \SI{5}{agents \per \meter^2}. 
The log residual error for a group of $n$ agents where the coordinates of each agent are $(x_i, y_i)$ is calculated as:
\begin{equation}
    \log \sum_{i=1}^n \big(y_i - (a_0 + a_1x_i) \big)^2,
\end{equation}
where $(a_0, a_1)$ are the learned least squares parameters over the $n$ agents.

For formation control, the experimental randomness model performs very closely to ideal communications, but simpler models show a wider variation of performance.
The log-scale formation residual error is shown in \fig{fig:form-error} and the time taken to converge to said formation is shown in \fig{fig:form-time}. 
Larger effects on performance are shown between the uniform and probabilistic disk models, where the lower communication range has a strong effect on performance accuracy, but the probabilistic model takes longer to converge as the agents lose and re-establish communications with nearby agents in the formation more frequently.

\section{Discussion and Future Work}
\label{sec:disc}
Our experiments demonstrate the importance of the network propagation model and schedule function on multi-agent task performance and agent scaling. 
There are two ways to view the results:
first, if the prior network model is full connectivity, introducing the more realistic model may seriously degrade performance and require adapting existing controllers.
On the other hand, the experimental randomness propagation model, as an improvement from the most conservative model using the lower bound of RF path loss, can represent an increase in potential task performance (e.g., average velocity while flocking as shown in Fig.~\ref{fig:flock_radius}). 
This change also motivates the need for refinement of network schedule functions, as allowing agents previously represented by a conservative unit-disc model to connect with more neighbors and move through space with more velocity presents a higher likelihood of losing connection with any given neighbor due to the physical dynamics.
\balance

The experiments presented here operate at a high level of abstraction, and do not necessarily point to specific improvements to be made in applications of networking and control. 
For example, this work has done little to optimize the scheduling function or inform how these communications models pose a risk for tasks where multi-hop connections are needed to send data to a central node, such as routing to a centralized controller during a search and rescue or exploration task. 
Creating methods for networks to add and drop members is also important, especially in the context of realistic communications.
\simname has been designed in order to help answer these, and other, questions in a more focused way in the future.

Although other communication methods used in swarms such as short-range infrared links or WiFi are not explored here, the RPC-based synchronization scheme that links the 6TiSCH simulator to the agent physical dynamics could be extended to include additional networking mediums in future work exploring multi-modal communication.

\section{Conclusion} 
\label{sec:conclusion}

This work presents a framework for understanding the performance and agent-count scaling of networked multi-agent robotic systems by more closely studying the effects of mobile RF communications. 
Although RF communication has been used for many experimental and simulated robotic tasks, improvements to the networking implementation using findings from the mobile wireless sensor networks community has been limited.
Using our simulator, we have shown that even relatively simple multi-agent control tasks become substantially more difficult when including realistic propagation models and scheduling functions, and these difficulties are compounded when increasing the number of agents.
The challenges we present in this paper will only be exacerbated when transitioning to the real world, and we hope \simname~will aid in designing new tools and techniques for networked, multi-agent control that scale to large, flexible, robot swarms.

\section*{Acknowledgments}
The authors would like to thank Brian Kilberg for his work on an earlier form of this simulator and the reviewers for their helpful feedback.

\clearpage    
\bibliographystyle{IEEEtran}
\bibliography{references}

\begin{thebibliography}{10}
\providecommand{\url}[1]{#1}
\csname url@samestyle\endcsname
\providecommand{\newblock}{\relax}
\providecommand{\bibinfo}[2]{#2}
\providecommand{\BIBentrySTDinterwordspacing}{\spaceskip=0pt\relax}
\providecommand{\BIBentryALTinterwordstretchfactor}{4}
\providecommand{\BIBentryALTinterwordspacing}{\spaceskip=\fontdimen2\font plus
\BIBentryALTinterwordstretchfactor\fontdimen3\font minus
  \fontdimen4\font\relax}
\providecommand{\BIBforeignlanguage}[2]{{%
\expandafter\ifx\csname l@#1\endcsname\relax
\typeout{** WARNING: IEEEtran.bst: No hyphenation pattern has been}%
\typeout{** loaded for the language `#1'. Using the pattern for}%
\typeout{** the default language instead.}%
\else
\language=\csname l@#1\endcsname
\fi
#2}}
\providecommand{\BIBdecl}{\relax}
\BIBdecl

\bibitem{brambilla2013swarm}
M.~Brambilla, E.~Ferrante, M.~Birattari, and M.~Dorigo, ``Swarm robotics: a
  review from the swarm engineering perspective,'' \emph{Swarm Intelligence},
  vol.~7, no.~1, pp. 1--41, 2013.

\bibitem{floreano2015science}
D.~Floreano and R.~J. Wood, ``Science, technology and the future of small
  autonomous drones,'' \emph{Nature}, vol. 521, no. 7553, pp. 460--466, 2015.

\bibitem{csahin2004swarm}
E.~{\c{S}}ahin, ``Swarm robotics: From sources of inspiration to domains of
  application,'' in \emph{International workshop on swarm robotics}.\hskip 1em
  plus 0.5em minus 0.4em\relax Springer, 2004, pp. 10--20.

\bibitem{egerstedt2001formation}
M.~Egerstedt and X.~Hu, ``Formation constrained multi-agent control,''
  \emph{IEEE transactions on robotics and automation}, vol.~17, no.~6, pp.
  947--951, 2001.

\bibitem{oh2015survey}
K.-K. Oh, M.-C. Park, and H.-S. Ahn, ``A survey of multi-agent formation
  control,'' \emph{Automatica}, vol.~53, pp. 424--440, 2015.

\bibitem{siljak2011decentralized}
D.~D. Siljak, \emph{Decentralized control of complex systems}.\hskip 1em plus
  0.5em minus 0.4em\relax Courier Corporation, 2011.

\bibitem{yick2008wireless}
J.~Yick, B.~Mukherjee, and D.~Ghosal, ``Wireless sensor network survey,''
  \emph{Computer networks}, vol.~52, no.~12, pp. 2292--2330, 2008.

\bibitem{akyildiz2010wireless}
I.~F. Akyildiz and M.~C. Vuran, \emph{Wireless sensor networks}.\hskip 1em plus
  0.5em minus 0.4em\relax John Wiley \& Sons, 2010, vol.~4.

\bibitem{murphy_disaster_2016}
\BIBentryALTinterwordspacing
R.~R. Murphy, S.~Tadokoro, and A.~Kleiner, ``\BIBforeignlanguage{en}{Disaster
  {Robotics}},'' in \emph{\BIBforeignlanguage{en}{Springer {Handbook} of
  {Robotics}}}, ser. Springer {Handbooks}, B.~Siciliano and O.~Khatib,
  Eds.\hskip 1em plus 0.5em minus 0.4em\relax Cham: Springer International
  Publishing, 2016, pp. 1577--1604. [Online]. Available:
  \url{https://doi.org/10.1007/978-3-319-32552-1_60}
\BIBentrySTDinterwordspacing

\bibitem{shakeri_design_2019}
R.~Shakeri, M.~A. Al-Garadi, A.~Badawy, A.~Mohamed, T.~Khattab, A.~K. Al-Ali,
  K.~A. Harras, and M.~Guizani, ``Design {Challenges} of {Multi}-{UAV}
  {Systems} in {Cyber}-{Physical} {Applications}: {A} {Comprehensive} {Survey}
  and {Future} {Directions},'' \emph{IEEE Communications Surveys Tutorials},
  vol.~21, no.~4, pp. 3340--3385, 2019.

\bibitem{drew2021multi}
D.~S. Drew, ``Multi-agent systems for search and rescue applications,''
  \emph{Current Robotics Reports}, pp. 1--12, 2021.

\bibitem{rouvcek2019darpa}
T.~Rou{\v{c}}ek, M.~Pecka, P.~{\v{C}}{\'\i}{\v{z}}ek,
  T.~Pet{\v{r}}{\'\i}{\v{c}}ek, J.~Bayer, V.~{\v{S}}alansk{\`y}, D.~He{\v{r}}t,
  M.~Petrl{\'\i}k, T.~B{\'a}{\v{c}}a, V.~Spurn{\`y} \emph{et~al.}, ``Darpa
  subterranean challenge: Multi-robotic exploration of underground
  environments,'' in \emph{International Conference on Modelling and Simulation
  for Autonomous Systesm}.\hskip 1em plus 0.5em minus 0.4em\relax Springer,
  2019, pp. 274--290.

\bibitem{municio2019simulating}
E.~Municio, G.~Daneels, M.~Vu{\v{c}}ini{\'c}, S.~Latr{\'e}, J.~Famaey,
  Y.~Tanaka, K.~Brun, K.~Muraoka, X.~Vilajosana, and T.~Watteyne, ``Simulating
  6tisch networks,'' \emph{Transactions on Emerging Telecommunications
  Technologies}, vol.~30, no.~3, p. e3494, 2019.

\bibitem{demir2019diva}
A.~K. Demir and S.~Bilgili, ``Diva: a distributed divergecast scheduling
  algorithm for ieee 802.15. 4e tsch networks,'' \emph{Wireless Networks},
  vol.~25, no.~2, pp. 625--635, 2019.

\bibitem{tanaka2020trace}
Y.~Tanaka, K.~Brun-Laguna, and T.~Watteyne, ``Trace-based simulation for
  6tisch,'' \emph{Internet Technology Letters}, vol.~3, no.~4, p. e162, 2020.

\bibitem{mehta2012mobility}
A.~M. Mehta, ``Mobility in wireless sensor networks,'' Ph.D. dissertation, UC
  Berkeley, 2012.

\bibitem{silva2014mobility}
R.~Silva, J.~S. Silva, and F.~Boavida, ``Mobility in wireless sensor
  networks--survey and proposal,'' \emph{Computer Communications}, vol.~52, pp.
  1--20, 2014.

\bibitem{ali2005mmac}
M.~Ali, T.~Suleman, and Z.~A. Uzmi, ``Mmac: A mobility-adaptive, collision-free
  mac protocol for wireless sensor networks,'' in \emph{IEEE International
  Performance, Computing, and Communications Conference}.\hskip 1em plus 0.5em
  minus 0.4em\relax IEEE, 2005, pp. 401--407.

\bibitem{khan2013collision}
B.~M. Khan and F.~H. Ali, ``Collision free mobility adaptive (cfma) mac for
  wireless sensor networks,'' \emph{Telecommunication Systems}, vol.~52, no.~4,
  pp. 2459--2474, 2013.

\bibitem{gu2012eswc}
Y.~Gu, Y.~Ji, J.~Li, and B.~Zhao, ``Eswc: Efficient scheduling for the mobile
  sink in wireless sensor networks with delay constraint,'' \emph{Transactions
  on Parallel and Distributed Systems}, vol.~24, no.~7, pp. 1310--1320, 2012.

\bibitem{vincze2005multi}
Z.~Vincze and R.~Vida, ``Multi-hop wireless sensor networks with mobile sink,''
  in \emph{Proceedings of the 2005 ACM conference on Emerging network
  experiment and technology}, 2005, pp. 302--303.

\bibitem{route-swarm}
R.~K. {Williams}, A.~{Gasparri}, and B.~{Krishnamachari}, ``Route swarm:
  Wireless network optimization through mobility,'' in \emph{IEEE International
  Conference on Intelligent Robots and Systems}, 2014, pp. 3775--3781.

\bibitem{robot-msg-ferrying}
S.~{Wang}, A.~{Gasparri}, and B.~{Krishnamachari}, ``Robotic message ferrying
  for wireless networks using coarse-grained backpressure control,'' \emph{IEEE
  Transactions on Mobile Computing}, vol.~16, no.~2, pp. 498--510, 2017.

\bibitem{balch1994communication}
T.~Balch and R.~C. Arkin, ``Communication in reactive multiagent robotic
  systems,'' \emph{Autonomous robots}, vol.~1, no.~1, pp. 27--52, 1994.

\bibitem{moreau2005stability}
L.~Moreau, ``Stability of multiagent systems with time-dependent communication
  links,'' \emph{IEEE Transactions on automatic control}, vol.~50, no.~2, pp.
  169--182, 2005.

\bibitem{zhang2018fully}
K.~Zhang, Z.~Yang, H.~Liu, T.~Zhang, and T.~Ba{\c{s}}ar, ``Fully decentralized
  multi-agent reinforcement learning with networked agents,'' \emph{arXiv
  preprint arXiv:1802.08757}, 2018.

\bibitem{arkin2002line}
R.~C. Arkin and J.~Diaz, ``Line-of-sight constrained exploration for reactive
  multiagent robotic teams,'' in \emph{International Workshop on Advanced
  Motion Control}.\hskip 1em plus 0.5em minus 0.4em\relax IEEE, 2002, pp.
  455--461.

\bibitem{zelazo2012rigidity}
D.~Zelazo, A.~Franchi, F.~Allg{\"o}wer, H.~H. B{\"u}lthoff, and P.~R. Giordano,
  ``Rigidity maintenance control for multi-robot systems,'' in \emph{Robotics:
  science and systems}, 2012, pp. 473--480.

\bibitem{de2006decentralized}
M.~C. De~Gennaro and A.~Jadbabaie, ``Decentralized control of connectivity for
  multi-agent systems,'' in \emph{IEEE Conference on Decision and
  Control}.\hskip 1em plus 0.5em minus 0.4em\relax IEEE, 2006, pp. 3628--3633.

\bibitem{yang2008multi}
P.~Yang, R.~A. Freeman, and K.~M. Lynch, ``Multi-agent coordination by
  decentralized estimation and control,'' \emph{IEEE Transactions on Automatic
  Control}, vol.~53, no.~11, pp. 2480--2496, 2008.

\bibitem{su2009flocking}
H.~Su, X.~Wang, and Z.~Lin, ``Flocking of multi-agents with a virtual leader,''
  \emph{IEEE transactions on automatic control}, vol.~54, no.~2, pp. 293--307,
  2009.

\bibitem{filippidis2012decentralized}
I.~Filippidis, D.~V. Dimarogonas, and K.~J. Kyriakopoulos, ``Decentralized
  multi-agent control from local ltl specifications,'' in \emph{IEEE Conference
  on Decision and Control}.\hskip 1em plus 0.5em minus 0.4em\relax IEEE, 2012,
  pp. 6235--6240.

\bibitem{jimenez2018decentralized}
A.~C. Jim{\'e}nez, V.~Garc{\'\i}a-D{\'\i}az, and S.~Bola{\~n}os, ``A
  decentralized framework for multi-agent robotic systems,'' \emph{Sensors},
  vol.~18, no.~2, p. 417, 2018.

\bibitem{Koenig-2004-394}
N.~Koenig and A.~Howard, ``Design and use paradigms for gazebo, an open-source
  multi-robot simulator,'' in \emph{IEEE Conference on Intelligent Robots and
  Systems}, Sendai, Japan, Sep 2004, pp. 2149--2154.

\bibitem{airsim2017fsr}
\BIBentryALTinterwordspacing
S.~Shah, D.~Dey, C.~Lovett, and A.~Kapoor, ``Airsim: High-fidelity visual and
  physical simulation for autonomous vehicles,'' in \emph{Field and Service
  Robotics}, 2017. [Online]. Available: \url{https://arxiv.org/abs/1705.05065}
\BIBentrySTDinterwordspacing

\bibitem{9199255}
A.~R. {Cheraghi}, K.~{Actun}, S.~{Shahzad}, and K.~{Graffi}, ``Swarm-sim: A 2d
  3d simulation core for swarm agents,'' in \emph{International Conference on
  Intelligent Robotic and Control Engineering}, 2020, pp. 1--10.

\bibitem{soria2020swarmlab}
E.~Soria, F.~Schiano, and D.~Floreano, ``Swarmlab: a matlab drone swarm
  simulator,'' 2020.

\bibitem{demarco2018}
K.~DeMarco, E.~Squires, M.~Day, and C.~Pippin, ``Simulating collaborative
  robots in a massive multi-agent game environment ({SCRIMMAGE}),'' in
  \emph{Int. Symp. on Distributed Autonomous Robotic Systems}, 2018.

\bibitem{watteyne2012openwsn}
T.~Watteyne, X.~Vilajosana, B.~Kerkez, F.~Chraim, K.~Weekly, Q.~Wang,
  S.~Glaser, and K.~Pister, ``Openwsn: a standards-based low-power wireless
  development environment,'' \emph{Transactions on Emerging Telecommunications
  Technologies}, vol.~23, no.~5, pp. 480--493, 2012.

\bibitem{robonetsim}
\BIBentryALTinterwordspacing
M.~Kudelski, L.~M. Gambardella, and G.~A. Di~Caro, ``Robonetsim: An integrated
  framework for multi-robot and network simulation,'' \emph{Robot. Auton.
  Syst.}, vol.~61, no.~5, p. 483–496, May 2013. [Online]. Available:
  \url{https://doi.org/10.1016/j.robot.2013.01.003}
\BIBentrySTDinterwordspacing

\bibitem{rosnetsim}
M.~Calvo-Fullana, D.~Mox, A.~Pyattaev, J.~Fink, V.~Kumar, and A.~Ribeiro,
  ``Ros-netsim: A framework for the integration of robotic and network
  simulators,'' \emph{Preprint}, 2020.

\bibitem{issariyakul2009introduction}
T.~Issariyakul and E.~Hossain, ``Introduction to network simulator 2 (ns2),''
  in \emph{Introduction to network simulator NS2}.\hskip 1em plus 0.5em minus
  0.4em\relax Springer, 2009, pp. 1--18.

\bibitem{Pinciroli:SI2012}
C.~Pinciroli, V.~Trianni, R.~O'Grady, G.~Pini, A.~Brutschy, M.~Brambilla,
  N.~Mathews, E.~Ferrante, G.~{Di Caro}, F.~Ducatelle, M.~Birattari, L.~M.
  Gambardella, and M.~Dorigo, ``{ARGoS}: a modular, parallel, multi-engine
  simulator for multi-robot systems,'' \emph{Swarm Intelligence}, vol.~6,
  no.~4, pp. 271--295, 2012.

\bibitem{quigley2009ros}
M.~Quigley, K.~Conley, B.~Gerkey, J.~Faust, T.~Foote, J.~Leibs, R.~Wheeler, and
  A.~Y. Ng, ``Ros: an open-source robot operating system,'' in \emph{ICRA
  workshop on open source software}, vol.~3, no. 3.2.\hskip 1em plus 0.5em
  minus 0.4em\relax Kobe, Japan, 2009, p.~5.

\bibitem{zavlanos2007flocking}
M.~M. Zavlanos, A.~Jadbabaie, and G.~J. Pappas, ``Flocking while preserving
  network connectivity,'' in \emph{IEEE Conference on Decision and
  Control}.\hskip 1em plus 0.5em minus 0.4em\relax IEEE, 2007, pp. 2919--2924.

\bibitem{yun1997line}
X.~Yun, G.~Alptekin, and O.~Albayrak, ``Line and circle formation of
  distributed physical mobile robots,'' \emph{Journal of Robotic Systems},
  vol.~14, no.~2, pp. 63--76, 1997.

\end{thebibliography}


\end{document}